\documentclass{article}
\usepackage[preprint]{neurips_2026}
\usepackage[utf8]{inputenc}
\usepackage[T1]{fontenc}
\usepackage{amsmath,amssymb,amsthm}
\usepackage{booktabs,array,graphicx,microtype,needspace}
\usepackage{xcolor}
\usepackage{url}
\usepackage[hidelinks]{hyperref}
\newcolumntype{L}[1]{>{\raggedright\arraybackslash}p{#1}}
\newtheorem{proposition}{Proposition}
\newcommand{\W}{\mathbf W}
\newcommand{\kk}{\mathbf k}
\newcommand{\vv}{\mathbf v}
\newcommand{\qq}{\mathbf q}
\newcommand{\I}{\mathbf I}
\newcommand{\gp}{g_{\mathrm{pred}}}
\newcommand{\gt}{g_{\mathrm{target}}}
\newcommand{\gz}{g_z}
\newcommand{\gb}{g_{\beta}}
\newcommand{\cv}{\mathrm{CV}}
\makeatletter
\renewcommand{\@noticestring}{Preprint.}
\makeatother
\title{Error Correction in Trained TTT Layers}
\author{Yuhe Hu \\ Duke University \\ \texttt{yh353@duke.edu}}
\begin{document}
\maketitle
\begin{abstract}
Test-time-training (TTT) layers update fast-weight memory while processing an input sequence. A gradient-based update, however, does not by itself establish how the memory revises an existing association. We study the structure and empirical role of prediction-derived updates in released TTT implementations. For dot-product objectives, we characterize the absence of a prediction-error residual and establish conditions under which inner-loop ascent and descent are equivalent after reparameterization. For the LayerNorm-based squared-error objective used by TTT-Linear and TTT-MLP, we derive the inner gradient and identify a prediction-derived component that cancels at the released initialization in the zero-epsilon limit. We then measure the full normalized prediction's contribution on three trained checkpoints. Removing this contribution while retaining the LayerNorm Jacobian gives a mean gradient norm ratio of $0.25$ to $0.32$ across the models. On $100{,}303$ tokens from \emph{War and Peace}, the same intervention with local per-token norm matching raises perplexity by $2.8$--$2.9\%$ on TTT-Linear-760M and 1.3B, whereas freezing all fast-weight updates raises it by factors of $4.2$--$4.5$. Separate synthetic tests illustrate the dependence of plain delta-rule overwriting on step size. The results distinguish a prediction-derived update from its behavioral value and show why implementation analysis and live interventions are both needed to assess test-time memory.
\end{abstract}

\section{Introduction}
\label{sec:intro}
Test-time-training (TTT) layers use fast weights as the recurrent state of a sequence model, updating them through an inner objective as the model processes its input \citep{sun2024ttt,zhang2025lact}. In key-value-binding implementations, learned projections produce keys, values, and queries. The inner update associates keys with values, and queries read from the resulting state. This design raises a specific question about test-time memory. When a later observation assigns a different value to an existing key, what enables the memory to revise the earlier association?

For a linear fast-weight memory, the distinction appears directly in the update rule. A dot-product objective produces an additive outer-product write. A squared-error objective subtracts the memory's current prediction before writing the new value. This is the classical delta rule, whose prediction-dependent subtraction can suppress a stale association under repeated writes to the same key \citep{widrow1960adaptive,schlag2021linear}. We study this mechanism without treating it as a definition of learning in general. Storing a new association and revising an existing one are different capabilities.

Canonical update equations do not fully describe released TTT systems. Nonlinear inner models, normalization, and learned update parameters can change the relationship between an objective and the update applied at inference. In particular, applying LayerNorm before a squared-error loss changes both the prediction-derived term and the term driven by the target. The remaining update can still depend on the current fast weights after the explicit prediction has been removed. This distinction matters when interpreting a gradient ablation as a test of error correction.

Gradient-sign experiments pose a related interpretive problem. \citet{liu2026secretly} report similar performance for ascent- and descent-trained dot-product TTT models and discuss how learned projections can absorb the sign change. We formalize a corresponding symmetry under explicit architectural and update conditions. The symmetry explains why parity alone need not distinguish competing accounts of the inner loop. It does not imply that reversing the gradient in an unchanged checkpoint preserves its outputs, nor that independently trained models must reach identical solutions.

Our empirical analysis focuses on TTT-Linear and TTT-MLP. We derive a decomposition of their released LayerNorm-based squared-error gradient, then inspect saved parameter statistics and live probes from three checkpoints. The measured intervention removes the full normalized prediction, including the LayerNorm bias, while retaining the normalization Jacobian. We state this operation explicitly because it differs from deleting only the component that vanishes for uniform gain.

We assess behavioral sensitivity using two trained TTT-Linear models. On approximately $100{,}000$ tokens from \emph{War and Peace}, removing the normalized prediction while matching each token's write norm at the current state raises perplexity by $2.8$--$2.9\%$. Freezing the entire fast-weight update raises perplexity by factors of $4.2$--$4.5$. These paired comparisons measure the effect on this text, not the general importance of revision. A separate synthetic task shows how plain delta-rule updates respond to conflicting associations, including at small effective steps. It does not measure the revision speed of the trained language models.

The paper contributes an analysis of gradient-sign symmetry for dot-product updates, a precise decomposition of the released LayerNorm-based squared-error update, and measurements connecting that decomposition to trained-model behavior. Together, these results support a practical sequence of checks. Inspect the implemented objective, measure its gradient on real activations, and test the behavior of interest directly.

\section{Inner objectives and prediction-derived updates}
\label{sec:analysis}
\subsection{Additive writes and the delta rule}
\label{sec:prelim}
Let $\W$ denote a fast-weight matrix, with column-vector keys $\kk$, values $\vv$, and queries $\qq$. For a linear inner model $f_\W(\kk)=\W\kk$, two common objectives give
\begin{align}
\ell_{\mathrm{dot}} &=-\langle\vv,\W\kk\rangle,
& \nabla_\W\ell_{\mathrm{dot}} &=-\vv\kk^\top, \label{eq:dot}\\
\ell_{\mathrm{mse}} &=\tfrac12\|\W\kk-\vv\|^2,
& \nabla_\W\ell_{\mathrm{mse}} &=(\W\kk-\vv)\kk^\top. \label{eq:mse}
\end{align}
The corresponding descent steps are
\begin{align}
\Delta\W_{\mathrm{dot}} &=\eta\vv\kk^\top,\\
\Delta\W_{\mathrm{mse}} &=\eta\vv\kk^\top-\eta(\W\kk)\kk^\top. \label{eq:delta}
\end{align}
The second term in Eq.~\eqref{eq:delta} removes part of the current prediction at the written key. For unit keys and $0<\eta\leq1$, repeated writes of the same target contract the residual in that direction. Additive writes retain the previous state and accumulate new associations. This comparison concerns the update rule, rather than every behavior that a network using the rule can express. Learned representations, readouts, and additional forgetting mechanisms can change how stored information is used.

For a nonlinear inner model $f_\W$, a dot-product loss still has output derivative $-\vv$, with no prediction-error residual. Its parameter gradient $-J_f^\top\vv$ can nevertheless depend on the current weights through $J_f$. Claims of state independence therefore apply to the linear update in Eq.~\eqref{eq:dot}, or to a linear final layer with its features held fixed, not to every parameter of a nonlinear TTT layer.

\Needspace{12\baselineskip}
\subsection{Gradient-sign symmetry}
\label{sec:gauge}
\begin{proposition}
\label{prop:gauge}
Consider $f(x)=\W_1h(x;\Theta)$ with a bias-free linear final fast-weight layer and inner loss $-\langle\vv,f(\kk)\rangle$. Suppose the network reads this branch as $Pf(\qq)$, and keys, values, queries, and inner step sizes are unchanged by the transformation below. Under simultaneous gradient updates, ascent from $(\W_1^{(0)},\Theta^{(0)},P)$ and descent from $(-\W_1^{(0)},\Theta^{(0)},-P)$ have negated $\W_1$ trajectories, identical $\Theta$ trajectories, and identical branch outputs.
\end{proposition}
The final-layer gradient does not depend on $\W_1$, while the gradient to $\Theta$ is odd in $\W_1$. Negating the initial final-layer weights and reversing the update sign therefore preserves the relationship at every step. Negating $P$ restores the readout. Appendix~\ref{app:gauge} gives the proof and the conditions needed for additional optimizer operations.

The result identifies corresponding parameterizations. It does not equate a frozen-checkpoint sign flip with retraining under the opposite sign. Nor does it force independently optimized models to converge to corresponding solutions. The practical implication is narrower. Where this symmetry applies, comparable performance after retraining is compatible with the same input-output computation and cannot alone establish the absence of memory.

LaCT and ViTTT use dot-product inner objectives in the configurations examined by \citet{liu2026secretly}. Their implementations require attention to branch structure. In the inspected ViTTT code, the ascent option reverses the SwiGLU gradients while leaving the convolution branch's update unchanged. A reparameterization must therefore be applied to the affected branch rather than to an entire shared output indiscriminately. Appendix~\ref{app:code} records the source locations.

\subsection{The released LayerNorm-based squared-error objective}
\label{sec:ln}
TTT-Linear and TTT-MLP do not use Eq.~\eqref{eq:mse} without modification. Their reference implementation forms a target $\vv'=\vv-\kk$ and applies LayerNorm to the inner prediction $z=f_\W(\kk)$ before the squared error \citep{sun2024ttt}. For head dimension $d$, define
\begin{equation}
\mu=\overline z,\qquad s^2=\overline{(z-\mu)^2}+\epsilon,\qquad
\hat z=(z-\mu\mathbf1)/s,\qquad y=\gamma\odot\hat z+\beta.
\end{equation}
The objective is $\ell=\tfrac12\|y-\vv'\|^2$. If
\begin{equation}
P_\epsilon(\hat z)=\I-\frac{\mathbf1\mathbf1^\top}{d}-\frac{\hat z\hat z^\top}{d},
\end{equation}
then its derivative with respect to $z$ is
\begin{align}
g &=\frac{1}{s}P_\epsilon(\hat z)[\gamma\odot(y-\vv')]
   =\gz+\gb+\gt, \label{eq:decomp}\\
\gz &=\frac{1}{s}P_\epsilon(\hat z)[\gamma^2\odot\hat z], \qquad
\gb =\frac{1}{s}P_\epsilon(\hat z)[\gamma\odot\beta], \notag\\
\gt &=-\frac{1}{s}P_\epsilon(\hat z)[\gamma\odot\vv']. \label{eq:parts}
\end{align}
Here $\gz$ comes from the normalized, gain-scaled prediction, and $\gb$ comes from its affine offset. We use $\gp=\gz+\gb$ for the full prediction-derived component. The remaining target component $\gt$ still depends on $\W$ through $\hat z$ and $s$. Removing $\gp$ therefore does not generally produce a state-independent additive memory.

\begin{proposition}
\label{prop:ln}
If $\gamma^2=c\mathbf1$, then
\begin{equation}
\gz=\frac{c\epsilon}{s^3}\hat z. \label{eq:epsilon}
\end{equation}
For nonzero prediction variance, $\gz$ vanishes when $\epsilon=0$. At the released initialization $\gamma=\mathbf1$ and $\beta=0$, the full prediction-derived component equals this epsilon residual. For nonuniform gain, its magnitude depends on both the gain pattern and the normalized activation.
\end{proposition}
Equation~\eqref{eq:epsilon} follows from $\overline{\hat z}=0$ and $\overline{\hat z^2}=1-\epsilon/s^2$. The reference implementation uses $\epsilon=10^{-6}$. Thus the cancellation is exact in the zero-epsilon limit, and the finite-epsilon term should be evaluated relative to the activation variance rather than assumed negligible for every input.

Channel variation of $\gamma^2$ is a useful parameter summary, but it is not a calibrated gate opening or a sufficient statistic for $\|\gz\|$. Two gain patterns with the same coefficient of variation can have different alignments with $\hat z$ and hence different projected components. Appendix~\ref{app:ln} gives an explicit example. Nonuniform gain permits $\gz$ to survive; it does not imply a monotone relationship between coefficient of variation and correction strength across arbitrary activations or checkpoints.

\section{Measurement protocol}
\label{sec:methods}
\subsection{Checkpoints and evaluation text}
We examine released TTT-Linear-760M, TTT-Linear-1.3B, and TTT-MLP-760M checkpoints trained on Pile-8k. Each has $24$ layers. The checkpoints are converted from the released JAX/EasyLM format to the reference PyTorch implementation. The conversion test compares synthetic checkpoints against a float64 NumPy transcription of the JAX forward computation, with deliberate mapping errors as negative controls. Appendix~\ref{app:tooling} describes this check and its limits.

The language-model evaluation uses the Project Gutenberg text of Tolstoy's \emph{War and Peace} in the Maude translation, tokenized with a Llama-2 tokenizer. Each ablation arm scores the same $49$ non-overlapping windows of $2048$ tokens, giving $100{,}303$ scored next-token positions. The fast-weight state is reset between windows. The text may overlap with pretraining, and this evaluation is not a held-out-corpus or explicit fact-revision benchmark. Pairing the inputs makes the interventions comparable on this text, but does not remove the limitation on generalization.

The saved TTT-Linear-760M gradient probe covers the first window, or $128$ mini-batches per layer. The standalone TTT-Linear-1.3B and TTT-MLP-760M probes cover four windows, or $512$ mini-batches per layer. These differences in coverage are retained in Table~\ref{tab:probe}. The language-model forward passes use float32, while the local gradient decomposition is recomputed in float64.

\subsection{Gradient decomposition and parameter summaries}
\label{sec:probe_method}
The stored runs use the harness option \texttt{drop\_pred}. It replaces the output residual $y-\vv'$ with $-\vv'$ before applying the LayerNorm Jacobian. In Eq.~\eqref{eq:decomp}, it retains $\gt$ and removes $\gp=\gz+\gb$. The alternative option \texttt{drop\_w} retains $\gb+\gt$ and removes only $\gz$. The results in this paper concern \texttt{drop\_pred}; we do not assume that the two interventions have the same effect because the learned $\beta$ may be nonzero.

For each call to the fused inner-gradient function, the probe records
\begin{equation}
R=\frac{\|g-\gt\|_F}{\|g\|_F+10^{-30}}
 =\frac{\|\gp\|_F}{\|g\|_F+10^{-30}}. \label{eq:ratio}
\end{equation}
The Frobenius norm is taken over the full tensor in that call, including the batch, head, token, and feature dimensions. We average these call-level ratios within each layer, then average the layer means. This is a norm ratio, not a fraction of explained variance or an allocation of causal importance. It can exceed one when components cancel. It is also not the ratio of the corresponding aggregated fast-weight updates, which additionally depends on keys and step sizes.

For each layer, the parameter summary $\cv(\gamma^2)$ is computed across channels within each head and then averaged across heads, using population standard deviations. We report its association with $R$ using Spearman correlation across layers. The $95\%$ percentile intervals use $10{,}000$ paired layer resamples and average ranks for ties, including the duplicated observations introduced by resampling. These intervals describe the layer association under that resampling scheme. They do not quantify uncertainty across independently trained models or evaluation corpora.

\subsection{Inference-time interventions}
\label{sec:arms}
We compare five arms on the two TTT-Linear checkpoints. The \emph{baseline} uses the released gradient $g$. The \emph{prediction-removed} arm uses $\gt$. The \emph{norm-matched} arm uses
\begin{equation}
\widetilde g_t={\gt}_t\frac{\|g_t\|_2}{\|{\gt}_t\|_2+10^{-30}}, \label{eq:normmatch}
\end{equation}
where the norms are taken over features separately for each token and head. The \emph{ascent} arm uses $-g$, and the \emph{frozen} arm uses zero inner gradients. All outer parameters remain fixed.

Norm matching compares $g_t$ and ${\gt}_t$ at the same current state within the modified trajectory. For a rank-one write $-\eta_t g_t\kk_t^\top$, this preserves the local Frobenius norm when the denominator is nonzero. It does not preserve the norm of the sum of writes within a chunk, whose cross terms depend on their directions, or the norm along the separately evolving baseline trajectory. The control tests whether the observed effect persists after compensating for local gradient magnitude.

The saved result files contain paired mean losses and summaries by context position. Their token-level standard errors do not account for sequence dependence, and grouping the same chunk position across windows does not provide independent document replicates. We therefore report the corpus-level effects descriptively without treating token-based or position-based errors as calibrated uncertainty estimates. A multi-document evaluation is needed for that purpose.

\section{Results}
\label{sec:results}
\subsection{Prediction-derived gradients in trained models}
\label{sec:probe}
The mean ratio $R$ is $0.265$ on TTT-Linear-760M, $0.251$ on TTT-Linear-1.3B, and $0.318$ on TTT-MLP-760M. Table~\ref{tab:probe} gives the probe coverage and correlations with gain variation. In all three checkpoints, layer~0 has both the largest gain-variation summary and the largest measured ratio. The association across the remaining layers is less consistent.

\begin{table}[t]
\centering\small
\caption{Live probes of the prediction-derived gradient. The ratio is defined in Eq.~\eqref{eq:ratio}. Chunks are counted per layer. Correlation intervals are exploratory paired layer-bootstrap intervals with average ranks for ties.}
\label{tab:probe}
\begin{tabular}{@{}lrrrc@{}}
\toprule
Checkpoint & Chunks & Mean $R$ & Median layer $\cv(\gamma^2)$ & $\rho$ [$95\%$ interval] \\
\midrule
TTT-Linear-760M & 128 & 0.265 & 0.504 & $0.490$ [$0.015,0.794$] \\
TTT-Linear-1.3B & 512 & 0.251 & 0.584 & $0.315$ [$-0.128,0.694$] \\
TTT-MLP-760M & 512 & 0.318 & 0.536 & $0.512$ [$0.124,0.769$] \\
\bottomrule
\end{tabular}
\end{table}

\begin{figure}[t]
\centering
\includegraphics[width=\linewidth]{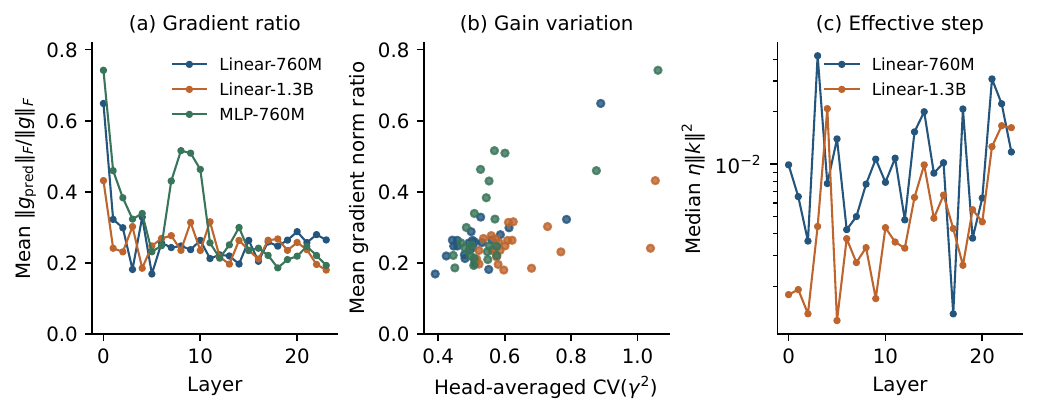}
\caption{Gradient probes from the saved runs. Each point is a layer. The left panel shows the mean norm ratio by depth. The center compares it with gain variation. The right shows per-layer medians of $\eta\|\kk\|^2$ for the two linear models. This last quantity is a reference scale from the plain delta rule, not a contraction factor for the LayerNorm objective.}
\label{fig:probe}
\end{figure}

For TTT-Linear-760M, Spearman $\rho=0.490$ with interval $[0.015,0.794]$. Removing the maximum-variation layer reduces it to $0.421$, with interval $[-0.072,0.757]$. The full-layer interval includes zero on TTT-Linear-1.3B. These measurements support using gain variation as a descriptive parameter summary, but not as a substitute for the live probe. The larger linear model has higher median gain variation and a slightly smaller mean $R$. The unequal probe coverage and activation dependence preclude attributing this difference to gain variation alone.

The linear-model probes also record the diagonal inner step factor multiplied by the squared key norm. Per-layer medians range from $0.00139$ to $0.04192$ on 760M and from $0.00127$ to $0.02086$ on 1.3B. These values include the learned token-position factor and learning-rate gate. For the plain delta rule, $\eta\|\kk\|^2$ directly determines contraction along the written key. For the released normalized objective, the Jacobian and activation statistics also enter, so the same scalar is not a complete description of update dynamics.

\subsection{Effects on language-model perplexity}
\label{sec:ladder}
Table~\ref{tab:ladder} reports the five-arm comparison. On TTT-Linear-760M, removing the full prediction increases perplexity from $12.77$ to $13.15$, or $2.93\%$. With local norm matching it reaches $13.13$, an increase of $2.83\%$. The corresponding increases on TTT-Linear-1.3B are $2.99\%$ and $2.85\%$. The effect persists after matching the local write norm, so a change in that norm alone does not explain the degradation.

\begin{table}[t]
\centering\small
\caption{Inference-time interventions on $100{,}303$ paired token positions from \emph{War and Peace}. The prediction-removed arm deletes both $\gz$ and $\gb$. Norm matching is local to each token and head. NLL differences are in nats per token.}
\label{tab:ladder}
\begin{tabular}{@{}lrrrr@{}}
\toprule
& \multicolumn{2}{c}{TTT-Linear-760M} & \multicolumn{2}{c}{TTT-Linear-1.3B} \\
\cmidrule(lr){2-3}\cmidrule(l){4-5}
Arm & Perplexity & $\Delta$NLL & Perplexity & $\Delta$NLL \\
\midrule
Baseline & 12.77 & 0 & 11.34 & 0 \\
Prediction removed & 13.15 & 0.02891 & 11.68 & 0.02943 \\
Prediction removed, norm matched & 13.13 & 0.02789 & 11.66 & 0.02811 \\
Ascent & 2595.69 & 5.31423 & 1530.27 & 4.90509 \\
Frozen & 54.18 & 1.44484 & 50.56 & 1.49505 \\
\bottomrule
\end{tabular}
\end{table}

\begin{figure}[t]
\centering
\includegraphics[width=\linewidth]{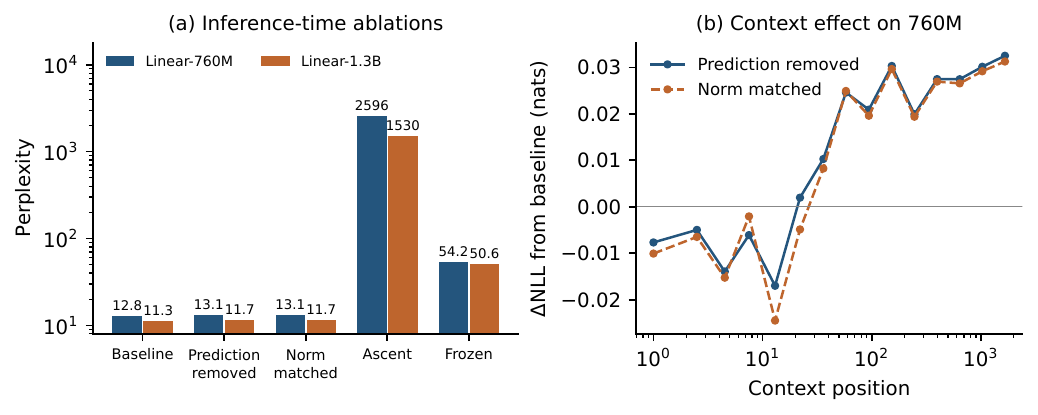}
\caption{Five inference-time interventions on the two linear checkpoints. Left, perplexity by arm. Right, the paired NLL difference by context position on 760M. Curves summarize the fixed evaluation text, with no independence assumption about adjacent tokens.}
\label{fig:ladder}
\end{figure}

Freezing the fast weights produces much larger changes, increasing perplexity by factors of $4.24$ and $4.46$. The models therefore depend strongly on their inner updates, while the tested prediction-removal intervention has a smaller effect on this text. This comparison is not a causal allocation of the inner loop's value. Perplexity is nonlinear, the components were trained together, and the retained target term remains state dependent.

The norm-matching multiplier has mean $1.017$ and median $1.012$ on 760M, and mean $1.015$ and median $1.010$ on 1.3B. These summaries show that rescaling is modest on average. They do not by themselves establish an angle between the components. Inferring orthogonality would require joint measurements of their inner products or consistent pointwise norms, rather than combining separately averaged ratios.

The prediction-removal penalty is small or negative near the start of the context and generally larger later in the window (Figure~\ref{fig:ladder}). This pattern is compatible with an accumulating effect of the intervention across chunk updates. It does not identify stale-binding interference as the cause, because the evaluation does not explicitly manipulate conflicting associations.

\subsection{What gradient ascent tests}
\label{sec:ascent}
Flipping the inner gradient at inference gives perplexities of $2595.69$ on 760M and $1530.27$ on 1.3B, factors of $203.2$ and $135.0$ above the corresponding baselines. This demonstrates sign sensitivity in these descent-trained checkpoints. It does not answer whether an outer loop could learn to use ascent, since the outer parameters are unchanged.

A separate matched-training experiment in the synthetic-task notebook addresses that narrower training question for a small plain-MSE model. An eight-class fast-weight classifier is trained from scratch under each sign convention for $600$ outer steps. Across three seeds, descent reaches $79.47\%\pm0.52\%$ accuracy and ascent reaches $70.38\%\pm0.52\%$, where the variation is the sample standard deviation across seeds. Both variants learn on this task, and descent performs better under this configuration. The experiment uses a plain squared-error objective and does not establish the corresponding gap for the released language models. Architecture and training details appear in Appendix~\ref{app:synthetic}.

\subsection{Overwriting in an idealized memory}
\label{sec:overwrite}
We use a synthetic task to make the difference between Eq.~\eqref{eq:dot} and Eq.~\eqref{eq:delta} explicit. A unit key in $d=128$ is associated first with an old random unit value and then with a new one. Each write uses a noisy normalized copy of the underlying key. Starting from $\W_0=0$, we query the underlying key and measure the cosine similarity to the new value minus that to the old value. Each cell contains $1000$ independent trials, with the same random draws used for the two rules.

\begin{figure}[t]
\centering
\includegraphics[width=\linewidth]{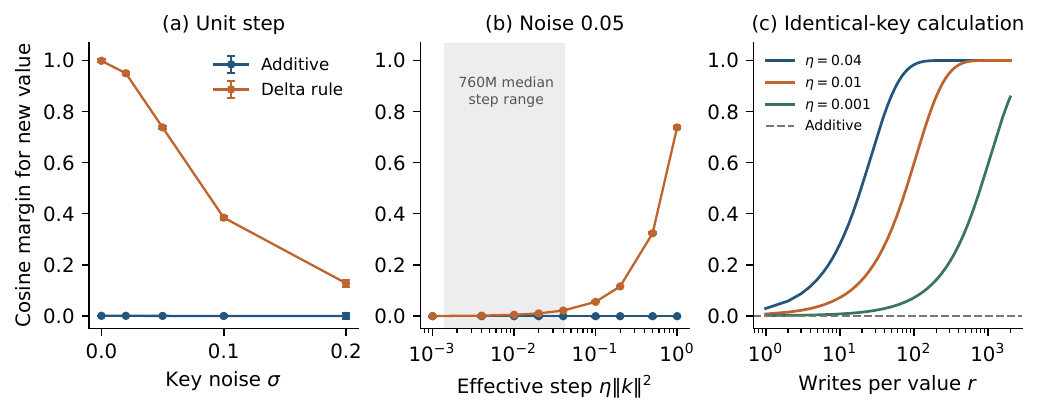}
\caption{Plain-rule overwrite behavior, separate from the trained language models. Panels (a) and (b) show the saved two-write simulation, independently checked using a closed-form calculation. Bars denote $95\%$ normal-approximation intervals over independent trials. The shaded range in (b) marks the measured linear-760M median step range only as a numerical reference. Panel (c) is an exact analytical calculation for identical unit keys and orthogonal values with $r$ old writes followed by $r$ new writes. It is not the noisy-key simulation or a trained-model measurement.}
\label{fig:overwrite}
\end{figure}

At unit step size, the delta-rule mean margins are $0.999$, $0.949$, $0.738$, $0.384$, and $0.128$ for key-noise levels $0$, $0.02$, $0.05$, $0.10$, and $0.20$. The additive mean margins remain within $0.001$ of zero. These values come from the stored NumPy sweep used in Figure~\ref{fig:overwrite}; the notebook's separate draw of the same task gives similar values. The delta rule preferentially retains the later association, while equal-weight additive writes have no systematic preference in this construction.

At fixed noise $0.05$, the single-write advantage decreases with the step size. This motivates an exact repeated-write calculation. For an identical unit key, orthogonal unit values, and $0<\eta\leq1$, let $q=(1-\eta)^r$. After $r$ old writes followed by $r$ new writes, the delta-rule readout is
\begin{equation}
\W_{2r}\kk=(1-q)(q\vv_{\mathrm{old}}+\vv_{\mathrm{new}}),
\qquad m_r=\frac{1-q}{\sqrt{1+q^2}}, \label{eq:overwrite}
\end{equation}
where $m_r$ is the cosine margin. Equal-count additive writes give margin zero. For small $\eta$, $q\approx e^{-\eta r}$, so the number of writes needed for a fixed margin grows approximately as $1/\eta$. The dependence on $\eta r$ is a small-step approximation, not an exact identity for all step sizes.

Equation~\eqref{eq:overwrite} explains slow replacement in this idealized rule. It does not turn the step statistics of Section~\ref{sec:probe} into a revision time for a trained TTT model. The language models use LayerNorm, biases, learned and changing representations, and chunked updates. An in-context benchmark with an explicit fact change is needed to measure their ability to replace a stale association.

\section{Related work}
\label{sec:related}
TTT-Linear and TTT-MLP use learned inner objectives and expressive recurrent states \citep{sun2024ttt}. LaCT and ViTTT extend the use of test-time updates across language and vision settings \citep{zhang2025lact,han2025vit3}. Titans and MIRAS develop related views of test-time memory and retention \citep{behrouz2025titans,behrouz2025miras}. Our analysis concerns the implemented objectives and checkpoints studied here. It does not cover every test-time adaptation method or end-to-end TTT trained through a final task loss \citep{tandon2025e2e}.

Additive fast-weight memories connect linear attention to recurrent computation \citep{katharopoulos2020transformers,schlag2021linear}. The delta rule addresses interference when keys are reused, and subsequent work develops parallel and gated implementations \citep{yang2024deltanet,yang2025gated}. The distinction between additive and corrective writes is established in that literature. Our contribution is to examine how a normalized inner objective changes the decomposition and how removing its prediction term affects trained checkpoints.

\citet{liu2026secretly} relate key-value-binding TTT to learned linear attention and use gradient-sign interventions in their empirical argument. We formalize a symmetry relevant to interpreting that intervention, while preserving the distinction between retraining and a frozen-checkpoint change. The result complements their sign-absorption explanation and does not address all of their empirical observations.

Work on test-time regression and mesa-optimization also connects learned sequence models to data-dependent inner updates \citep{wang2025testtime,vonoswald2023mesa}. Our measurements follow that mechanistic approach while separating parameter summaries, local gradient ratios, and downstream interventions. Those quantities answer different questions and should not be substituted for one another.

\section{Discussion and limitations}
\label{sec:discussion}
The analysis gives a precise reason to inspect an implementation before interpreting its inner loop. Dot-product and squared-error objectives differ in whether they contain an explicit prediction-error residual. LayerNorm then changes the squared-error gradient enough that removing the prediction is no longer equivalent to replacing the update with a state-independent additive write. A claim about error correction must specify both the algebraic component and the behavior being measured.

The trained-model evidence is consistent across the two linear checkpoints. Prediction removal causes a modest increase in perplexity on the evaluated text, including under local norm matching, while freezing the fast weights causes a much larger increase. The MLP probe confirms a nonzero prediction-derived component in that model family, but no MLP perplexity ablation is reported. These findings establish sensitivity to the tested component. They do not show that the remaining update is incapable of revising information, or that the deleted component is the only mechanism capable of doing so.

Several limitations constrain the interpretation. The evaluation contains one public-domain work and may overlap with pretraining. It does not sample independent documents or directly test contradictory facts. The probes cover one or four windows, and their ratios depend on activations as well as parameters. Layer-bootstrap intervals do not account for dependence between layers or variability across training runs. The main intervention deletes the affine offset contribution together with the normalized prediction, so its effect cannot be assigned to $\gz$ alone. Local norm matching does not control the summed chunk-update norm or align entire trajectories.

The overwrite task supplies a controlled example of the plain delta rule, but it is not a substitute for an overwrite test in a trained language model. Likewise, the small matched-training experiment does not determine whether large normalized TTT models trained under ascent would match descent. The conversion test checks mapping and numerical consistency on synthetic checkpoints, not an independent full-scale evaluation in the original JAX runtime.

A useful next evaluation would combine a held-out multi-document corpus, direct measurement of component angles, separate ablations of $\gz$ and $\gb$, and a controlled in-context revision task. The present evidence supports a narrower conclusion. Prediction-derived updates are measurable and behaviorally relevant in the checkpoints examined here, while their role in revising semantic associations remains to be tested directly.

\begin{ack}
We thank the reviewers of the NeurIPS 2026 Workshop on Test-Time Continual Learning for comments that motivated the norm-matched control, replication, and step-size analysis.
\end{ack}

\bibliographystyle{plainnat}
\bibliography{references}

\appendix
\section{Proof and scope of the sign symmetry}
\label{app:gauge}
For $\ell=-\langle\vv,\W_1h(\kk;\Theta)\rangle$, the gradients are
\begin{equation}
\nabla_{\W_1}\ell=-\vv h^\top,\qquad
\nabla_\Theta\ell=-J_h^\top\W_1^\top\vv.
\end{equation}
Write $A$ for the ascent trajectory and $D$ for the descent trajectory. Suppose $\W_{1,D}=-\W_{1,A}$ and $\Theta_D=\Theta_A$ at one step. Their $\W_1$ gradients coincide, while their $\Theta$ gradients have opposite signs. With a shared step size,
\begin{align}
\W_{1,D}^{+}&=-\W_{1,A}-\eta\nabla_{\W_1}\ell_A=-\W_{1,A}^{+},\\
\Theta_D^{+}&=\Theta_A+\eta\nabla_\Theta\ell_A=\Theta_A^{+}.
\end{align}
Induction gives the state relation for the whole inner sequence. Since $P_D=-P_A$, their readouts agree. This proof also applies to a batch sum of the dot-product objective with the same inputs and step factors.

Additional operations require corresponding symmetry conditions. A gradient transform $U$ must obey $U(-G)=-U(G)$ where an odd transformation is needed. Zero-initialized linear momentum preserves the relation with correspondingly transformed momentum states. The map $G\mapsto G/(\|G\|+\epsilon)$ is odd, not even. Norm-based weight normalization is also odd in the weights when its scale is unchanged. These properties should be checked for the actual optimizer and branch composition rather than inferred from an optimizer name.

The proposition assumes a bias-free final layer. A fixed, untransformed additive branch can obstruct the output sign transformation. A trainable final bias may admit an extended transformation if the bias and its update are also negated, so the existence of any bias is not by itself a universal impossibility result. Shared projections, nonlinear readouts, or sign-dependent step sizes likewise require separate analysis.

A local float64 check using a nonlinear hidden map, $11$ inner steps, and matched sign-transformed parameters gives zero maximum absolute output difference. This validates the algebraic construction; it is not a claim of bitwise equivalence for every released kernel, precision, compiler, or optimizer configuration.

\section{LayerNorm derivation and numerical checks}
\label{app:ln}
Differentiating $\hat z=(z-\mu\mathbf1)/s$ yields the Jacobian $P_\epsilon(\hat z)/s$. Expanding the residual $y-\vv'$ gives Eq.~\eqref{eq:decomp}. With $\gamma^2=c\mathbf1$,
\begin{equation}
P_\epsilon(\hat z)[c\hat z]
=c\hat z-c\overline{\hat z}\mathbf1-c\hat z\overline{\hat z^2}
=\frac{c\epsilon}{s^2}\hat z,
\end{equation}
which proves Proposition~\ref{prop:ln}. When $\epsilon>0$, $P_\epsilon$ is not an exact orthogonal projection in the normalized-prediction direction. The zero-variance case must not be analyzed by setting both $s$ and $\epsilon$ to zero.

Gain variation alone cannot specify the surviving prediction component. Take $\epsilon=0$, $s=1$, and $\hat z=(1,-1,1,-1)$. The patterns $\gamma_A^2=(1,2,1,2)$ and $\gamma_B^2=(1,1,2,2)$ have the same mean, variance, and coefficient of variation. Nevertheless,
\begin{equation}
P_0(\hat z)[\gamma_A^2\odot\hat z]=0,\qquad
P_0(\hat z)[\gamma_B^2\odot\hat z]=(-\tfrac12,\tfrac12,\tfrac12,-\tfrac12).
\end{equation}
An empirical sweep that changes the gain pattern in one prescribed way may show a monotone trend. That trend is not a general theorem in $\cv(\gamma^2)$.

The reference function \texttt{ln\_fused\_l2\_bwd} agrees with automatic differentiation of $\tfrac12\|y-\vv'\|^2$ in a float64 random-input check to maximum absolute error $1.6\times10^{-15}$. The same check verifies that \texttt{drop\_pred} removes $\gz+\gb$, that \texttt{drop\_w} removes $\gz$, and that token-wise norm matching preserves the feature-vector norm. Tests also verify the finite-epsilon residual and the equal-variation counterexample above.

These checks support the implementation-level definitions. They do not establish that $\gt$ has no corrective behavior. Its dependence on the current prediction through the Jacobian remains present, and further parameter dependence appears when $g$ is propagated through a nonlinear inner model.

\section{Code and reproducibility}
\label{app:code}
\subsection{Source locations}
The inspected sources are listed below. The TTT validation kit includes a vendored copy of the reference PyTorch model, a checkpoint converter, the ablation functions, and scripts for probing and scoring. The released run scripts select \texttt{drop\_pred} and token-wise norm matching explicitly.
\begin{center}\small
\begin{tabular}{@{}L{0.51\linewidth}L{0.39\linewidth}@{}}
\toprule
Repository & Revision or source identity \\
\midrule
\texttt{JunchenLiu77/LaCT} & \texttt{1c2109fef1cf} \\
\texttt{JunchenLiu77/ViTTT} & \texttt{64cba5525a66} \\
\texttt{test-time-training/ttt-lm-pytorch} & \texttt{cd831db10c8c} \\
\texttt{test-time-training/ttt-lm-jax} & \texttt{6f529b124c7f} \\
\bottomrule
\end{tabular}
\end{center}
The LaCT source routes the selected loss through its fast-weight update implementation. In ViTTT, \path{vittt/models/ttt_block.py} computes the dot-product output derivative as a scaled negative value tensor. The \texttt{ga\_dot\_product} option negates the SwiGLU gradients. The depthwise-convolution update has no corresponding sign branch. These code-level observations establish the selected objective and intervention, rather than a universal claim about every configuration in either repository.

For TTT-Linear, \texttt{ttt.py} forms $z$ from the current linear fast weights and bias, then calls \texttt{ln\_fused\_l2\_bwd} with target $\vv-\kk$. TTT-MLP calls the same function on its final inner prediction and propagates the derivative through its hidden layer. The harness changes this module-level function, so the intervention reaches both the output computation and the carried fast-weight state in the reference implementation. The baseline reinstalls the original function. A small-model check confirms identical baseline outputs, and the frozen arm leaves the carried fast weights unchanged across chunks.

\subsection{Conversion and saved runs}
\label{app:tooling}
The converter reads the EasyLM streaming-msgpack checkpoint format, decodes bfloat16 payloads, transposes dense kernels, maps head-stacked parameters, and reconciles convolution and rotary-position layouts. Its test suite constructs byte-format-compatible synthetic checkpoints and compares the converted float32 PyTorch forward pass with a float64 NumPy transcription operating on the original parameter layout. The maximum relative error in the local rerun is $6.3\times10^{-7}$. All seven deliberately incorrect mappings are detected. This checks the converter and reference transcription on those fixtures, not equivalence against an independently executed full-scale JAX checkpoint.

The saved perplexity files are \texttt{ablation\_results\_linear760m.json} and \texttt{ablation\_results\_linear13b.json}. Their \texttt{per\_arm} and \texttt{deltas\_vs\_descent} fields supply Table~\ref{tab:ladder}. The first file also supplies the 760M gradient probe. The standalone files \texttt{probe\_linear-1.3b.json} and \texttt{probe\_mlp-760m.json} supply the other two probes. The 1.3B ablation file contains an additional one-window probe with mean ratio $0.250$, close to the four-window value $0.251$ reported in the main table.

The saved run configuration uses context length $2048$, mini-batch size $16$, float32 model execution, and a Llama-2 tokenizer. The Modal environment pins PyTorch~2.4.1 and Transformers~4.44.2. Each arm resets the fast-weight cache for each window and scores identical token positions. The corpus is downloaded by the run script from the public text mirror linked in the repository. The tables can be recomputed from the saved outputs.

The live ratio in Eq.~\eqref{eq:ratio} is recomputed in float64 before aggregation to reduce cancellation error. The norm-matching log instead summarizes token/head ratios per call. The two summaries use different aggregation operations, which is another reason not to infer component angles from their means. The code's historical arm names \texttt{hebbian} and \texttt{hebbian\_nm} refer to prediction removal and its norm-matched variant in the tables here. They should not be interpreted as proof of a state-independent Hebbian update.

\subsection{Available artifacts}
The existing experiment repository is \url{https://github.com/aliceyh7/ttt-error-correction}. It contains the conversion and intervention code and saved language-model and overwrite outputs. The ancillary material accompanying this source includes the table-recomputation script, corrected tied-rank bootstrap, figure-generation script, numerical checks, and the saved data used by them. It also includes the conversion-test log and the synthetic-classifier notebook outputs used for the matched-training result. The new analysis scripts do not overwrite the historical runs.

\section{Synthetic tasks and repeated writes}
\label{app:synthetic}
\subsection{Matched training}
The synthetic classifier has hidden dimension $48$ and eight classes. Each episode draws a Gaussian prototype for each class, samples $12$ labeled context examples and one query, and adds Gaussian noise with standard deviation $0.12$ to their prototypes. Learned projections produce normalized keys, value vectors from one-hot labels, and output logits from the fast-weight readout. Fast weights start at zero for each episode. The inner rule uses plain squared error with fixed step $0.35$ and the selected sign.

Each variant is trained with AdamW at learning rate $5\times10^{-4}$, batch size $96$, gradient clipping at norm $1$, and $600$ outer steps. The seeds are $0$, $1$, and $2$. Final evaluation uses $40$ fresh batches per seed. The saved notebook gives descent accuracies of $79.45\%$, $78.96\%$, and $80.00\%$ and ascent accuracies of $70.94\%$, $70.31\%$, and $69.90\%$ for the three seeds. Mean losses are $0.762$ and $1.011$, respectively. The ancillary notebook record preserves the per-seed values. These runs are separate from the language-model evaluation.

The comparison uses the same task and hyperparameters under each sign. It is one controlled configuration rather than a search over the best achievable model for each update rule.

\subsection{Two-write simulation}
The stored NumPy sweep draws an underlying unit key $\kk$, noisy normalized keys $\kk_i=\mathrm{normalize}(\kk+\sigma\xi_i)$ with independent $\xi_i\sim N(0,\I)$, and random unit values $\vv_{\mathrm{old}}$ and $\vv_{\mathrm{new}}$ in $d=128$. It starts from $\W_0=0$ and applies one old write and one new write. For the additive rule,
\begin{equation}
\W_2\kk=\eta\big[\vv_{\mathrm{old}}\langle\kk_1,\kk\rangle+
\vv_{\mathrm{new}}\langle\kk_2,\kk\rangle\big].
\end{equation}
For the delta rule,
\begin{equation}
\W_2\kk=\eta\vv_{\mathrm{old}}\langle\kk_1,\kk\rangle+
\eta\big[\vv_{\mathrm{new}}-\eta\vv_{\mathrm{old}}\langle\kk_1,\kk_2\rangle\big]
\langle\kk_2,\kk\rangle.
\end{equation}
Recomputing these expressions with the saved script's random-number order reproduces all $28$ cells of the two-write sweeps to maximum mean-margin difference $1.2\times10^{-16}$. Figure~\ref{fig:overwrite} uses those saved cell means and standard errors. We do not use the separately archived repeated-noisy-write results, whose complete generation procedure is not included in the released sweep script.

\subsection{Repeated identical keys}
For a unit key and plain MSE, a descent step updates its readout $u=\W\kk$ as $u^+=(1-\eta)u+\eta\vv$. Starting from zero, $r$ old writes give $(1-q)\vv_{\mathrm{old}}$, with $q=(1-\eta)^r$. A further $r$ new writes give Eq.~\eqref{eq:overwrite}. Orthogonality of the two unit values gives the stated cosine margin. The $\eta r$ approximation follows from $\log(1-\eta)=-\eta+O(\eta^2)$.

For nonunit keys, the contraction factor is $1-\eta\|\kk\|^2$, subject to the appropriate stability range. Under ascent the corresponding factor is $1+\eta\|\kk\|^2$, so the simple recurrence expands the old residual. These equations concern the plain linear rule. They should not be used as exact recursions for the LayerNorm objective.

The reference TTT implementation also evaluates within-chunk gradients at the chunk-start fast weights. A sequence of token-level contractions therefore cannot be identified directly with a sequence of chunked updates. Our synthetic comparison uses per-write updates, and our trained-model measurements use the released chunked implementation.
\end{document}